\documentclass[10pt,twocolumn,letterpaper]{article}

\usepackage{cvpr}              

\usepackage{graphicx}
\usepackage{amsmath}
\usepackage{amssymb}
\usepackage{booktabs}
\usepackage{xcolor}
\usepackage{MnSymbol,wasysym}
\usepackage{comment}
\usepackage{multicol}
\usepackage{multirow}
\usepackage{ctable}
\usepackage[flushleft]{threeparttable}
\usepackage{booktabs}

\usepackage[pagebackref,breaklinks,colorlinks]{hyperref}

\usepackage[capitalize]{cleveref}
\crefname{section}{Sec.}{Secs.}
\Crefname{section}{Section}{Sections}
\Crefname{table}{Table}{Tables}
\crefname{table}{Tab.}{Tabs.}

\def\cvprPaperID{9258} 
\def\confName{CVPR}
\def\confYear{2022}

\begin{document}

\title{Global Interaction Modelling in Vision Transformer via Super Tokens}

\author{Ammarah Farooq\\
{\tt\small ammarah.farooq@surrey.ac.uk}
\and
Muhammad Awais\\
{\tt\small m.a.rana@surrey.ac.uk}
\and
Sara Ahmed\\
{\tt\small  sara.atito@gmail.com}
\and
Josef Kittler\\
{\tt \small j.kittler@surrey.ac.uk}\\
CVSSP, University of Surrey
}
\maketitle

\begin{abstract}
   With the popularity of Transformer architectures in computer vision, the research focus has shifted towards developing computationally efficient designs. Window-based local attention is one of the major techniques being adopted in recent works. These methods begin with very small patch size and small embedding dimensions and then perform strided convolution (patch merging) in order to reduce the feature map size and increase embedding dimensions, hence, forming a pyramidal Convolutional Neural Network (CNN) like design. In this work, we investigate local and global information modelling in transformers by presenting a novel isotropic architecture that adopts local windows and special tokens, called Super tokens, for self-attention. Specifically, a single Super token is assigned to each image window which captures the rich local details for that window. These tokens are then employed for cross-window communication and global representation learning. Hence, most of the learning is independent of the image patches $(N)$ in the higher layers, and the class embedding is learned solely based on the Super tokens $(N/M^2)$ where $M^2$ is the window size. In standard image classification on Imagenet-1K, the proposed Super tokens based transformer (STT-S25) achieves 83.5\% accuracy which is equivalent to Swin transformer (Swin-B) with circa half the number of parameters (49M) and double the inference time throughput. The proposed Super token transformer offers a lightweight and promising backbone for visual recognition tasks.
\end{abstract}

\begin{figure}
\centering
        \includegraphics[width=\linewidth]{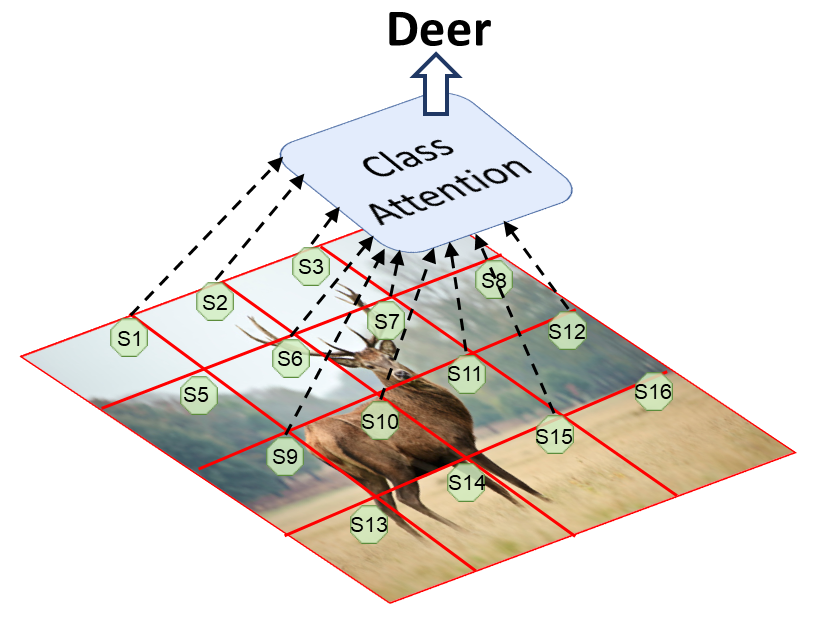}
        {\caption{Learning image class embedding from window Super tokens. The main idea is to learn a rich Super token for each window which is responsible for local and global information modelling. This is achieved by learning local context from the image window, preserving meaningful cross-window connections and global scale information of the objects.
        }
    \label{fig-illustration}}
\end{figure}
\section{Introduction}
\label{sec:intro}
Transformers~\cite{vaswani2017attention, devlin2018bert} have shown a remarkable improvement in performance for Natural Language Processing (NLP). This led to a profound interest in transformers in computer vision (CV)~\cite{dosovitskiy2020image, touvron2021training, liu2021swin, wang2021pyramid}. Transformers have achieved state-of-the-art results in multiple fields, e.g., speech~\cite{dong2018speech,gulati2020conformer}, NLP~\cite{devlin2018bert}, Self-Supervised Learning (SSL) in Computer Vision (CV)~\cite{caron2021emerging,atito2021sit}, multi-modal analysis~\cite{li2020oscar} etc. demonstrating their ability as a universal function approximator with the reuse of a simple multi-head attention block and a pointwise convolution (feed forward) network. The core advantage of transformers is their ability to model long-range/global (sequential/spatial) relations and to attend the relevant parts of the input for a given task. However, the need for large-scale image databases and resolutions, limited computing resources, and a greener/climate positive AI focus motivate the design of more efficient vision transformers.

The main computational overhead in a transformer block is the Multi-head Self Attention (MSA) which scales quadratically with the number of input data tokens. In CV, the number of data tokens is fairly high, due to the 2D spatial nature of the input. Several techniques have been proposed to address the computational complexity of transformers~\cite{touvron2021going, liu2021swin,chu2021twins,chen2021regionvit,fang2021msg, wang2021pyramid,kitaev2020reformer,yuan2021tokens,han2021transformer, el2021xcit}. Window-based self-attention (WMSA) has become one of the most popular mechanisms to reduce computational complexity of the self-attention operation~\cite{liu2021swin,chu2021twins,wang2021pyramid}. It constraints the MSA to local windows, consisting of a sub-set of data tokens. Later on, a cross-window interaction mechanism is utilised to share information between neighbouring windows. The use of windows allows a smaller patch size, which is beneficial to performance, but computationally expensive. To deal with this, the works in~\cite{liu2021swin,wang2021pyramid} proposed CNN-like pyramidal architecture for vision transformers. These hierarchical approaches use a smaller embedding size in initial blocks and do strided convolution (token merging) to change the feature resolution and increase the embedding dimension in later blocks. However, one main limitation of such designs is the interpretability in regard to the attention visualisation for decision making. This aspect is compromised due to the dilution of attention by the strided convolution (token merging) operation.

In this work, we focus on the problem of efficient transformer architecture design. We adopt window-based local self-attention. As depicted in Figure~\ref{fig-illustration}, we partition the image into spatial windows. Each window contains $M \times M$ data tokens. However, as we lose the global representation ability of transformers, we require an efficient way to model global information and long range dependencies. Different from existing mechanisms, to share information across different local windows, we deploy a learnable token called Super token (SupTk). Super tokens are randomly initialised and learnt along the corresponding window data tokens. Each Super token summarises the visual contents of all the data tokens within a window. A Super Token Mixer (STM) module then consolidates the information across all the Super tokens. After communicating global information with other Super tokens, each Super token goes back to its window, and updates the corresponding local tokens with respect to the global context. For example, the deer in Figure~\ref{fig-illustration} is distributed over windows with Super tokens s2, s3, s6, s7, s10, s11, s13, s14, s15. Once the local representation of face is built in the token s7, we would like to update it with respect to the rest of the Super tokens so that we can model strong links between the Super tokens on deer, and ignore the rest as background. 

With the brief idea of the proposed Super tokens based image transformer mentioned above, we essentially try to investigate the following research questions. Can we encapsulate image data tokens in the form of Super tokens,  summarising the image content, and in this way linking to the global context while maintaining local details? Are the summarised Super tokens enough for class prediction? How to model cross-window interactions with Super tokens for efficient learning? Do Super tokens help with the interpretability of the results? Does the proposed design offer any computational benefit? 

To summarise, our key contributions in this work are as follows:
\begin{enumerate}
    \item We propose a new isotropic architecture for vision transformer, based on local window attention and a global information consolidation by Super token interaction, to learn image representation. The proposed Super tokens do cross-window communication as well as model global information across the image.  
    \item The proposed Super Token Mixer is based on residual block design with simple point-wise and depth-wise convolutions to fuse the global information. It alleviates the need for expensive global self-attention across image patches ($N$). It is independent of any data token. By effectively reducing the number of tokens to ($N/{M^2}$), where $M \times M$ is the number of data tokens per window, it significantly improves the memory consumption.
    \item The proposed Super tokens based transformer (STT-S25) achieves 83.5\% Top-1 accuracy on the Imagenet-1K classification benchmark, which is equivalent to Swin base transformer~\cite{liu2021swin} (Swin-B) with about half the number of parameters (49M). The experimental results also verify the computational benefits of the proposed design. 
\end{enumerate}

\begin{figure*}
\centering
        \includegraphics[width=\textwidth]{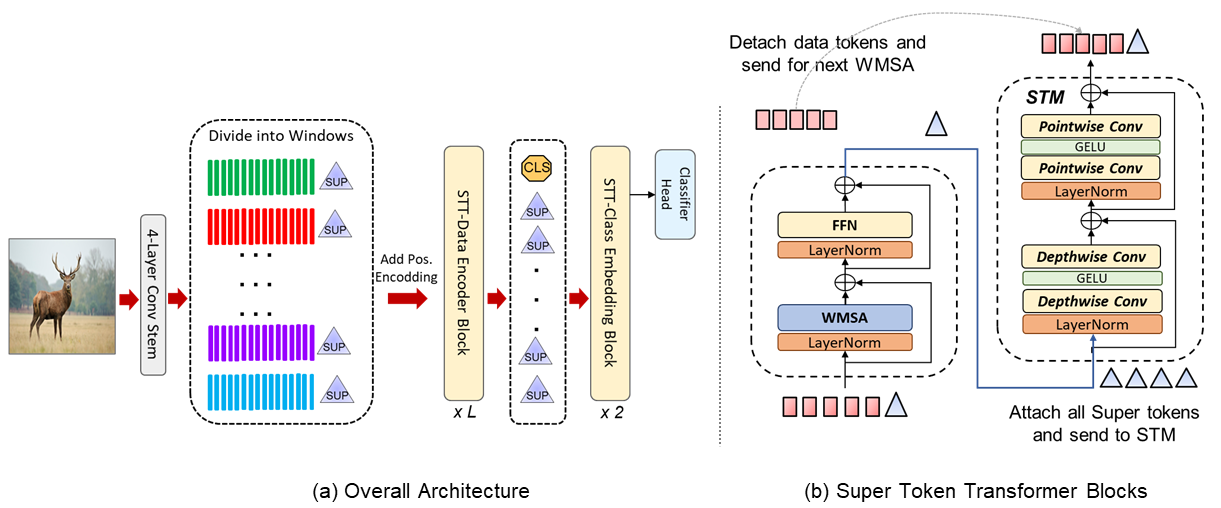}
        {\caption{(a) The overall architecture of the Super Token Transformer (STT). The data encoder blocks embed the visual content through an attention mechanism into Super tokens (SUP), and a class embedding block learns the final classification (CLS) token from the Super tokens. (b) Architecture of the STT-Data encoder blocks. First block processes all tokens with local-windowed self-attention. Next, the Super Token Mixer (STM) block consolidates the information at a global scale via all Super tokens.}
    \label{block-diagram}}
\end{figure*}

\section{Related Work}
\noindent Our work deals with the architectural design component of the vision transformers. Therefore, we discuss the three recent trends as related work.

\bigskip
\noindent \textbf{Vision Transformers.} The vanilla image transformers strictly follow the MSA-FFN style data encoder~\cite{vaswani2017attention} with MSA is formulated as:
\begin{center}
$MSA(Q, K, V ) = Concat(head_1, \cdots, head_h)W^O$
\end{center}
where $head_i = Attention(QW^{Qi}, KW^{Ki}, VW^{Vi})$ and 
\begin{center}
    $Attention(Q, K, V ) = softmax(QK^{T}/\sqrt{d_k})V$
\end{center}
Query ($Q$), Key ($K$) and Value ($V$) are the three linear projections of the input matrix and $d_k$ represents the feature dimension of $K$. Attention is calculated between $Q$ and $K$ and used to scale $V$. The first encoder block receives a set of feature maps as input coming from a convolutional block (sub-optimal case is using a single CNN layer as in~\cite{dosovitskiy2020image}). The convolutional block takes an input image $\mathbf{x} \in \mathbb{R}^{H \times W \times C}$ and converts it to feature maps of size $\mathbf{{x^f}} \in \mathbb{R}^{\sqrt{n} \times \sqrt{n} \times D}$, where, $H$, $W$, and $C$ are height, width and channels of the input image, $n$ is the total number of spatial locations in the feature maps and $D$ is the number of feature map channels. Each spatial location in the input feature maps is considered as an input data token to transformer, consisting of $n$ tokens in total. Both the input and output of the transformer have the same dimensions $\mathbf{x^f, y} \in \mathbb{R}^{n \times D}$. The earlier works ViT~\cite{dosovitskiy2020image}, Deit~\cite{touvron2021training} and CaiT~\cite{touvron2021going} followed this design. The Deit~\cite{han2021transformer} introduced distillation from the CNN teacher network while the CaiT~\cite{touvron2021going} proposed a layer scaling operation to aid with increasing depth of the transformers.

\bigskip
\noindent \textbf{Hierarchical Designs.} As described in the introduction, the hierarchical designs employ local-windowed attention and a cross-window communication strategy. The Swin transformer~\cite{liu2021swin} used a shifted window mechanism. Although the design served as an excellent backbone for the following works~\cite{chu2021twins,fang2021msg,chen2021regionvit,zhang2021aggregating,dong2021cswin}, the shifted window communication had an overhead of careful cyclic shift and padding. Shortly after, the work~\cite{chu2021twins} proposed two models, namely Twins-PCPVT and Twins-SVT. The later one introduced interleaved local-global style design. The global sub-sampled attention (GSA) module used strided convolution function to summarise the local tokens and apply MSA for global interaction. The MSGTr~\cite{fang2021msg} and RegionViT~\cite{chen2021regionvit} proposed a similar idea of interacting with a regional level representative tokens. We discuss these works in more detail in the next section to clarify the design differences.

\bigskip
\noindent \textbf{Mixer Networks.} Following the main idea of mixing information across channels and spatial locations through shared point-wise and depth-wise convolutions respectively, a cohort of works~\cite{tolstikhin2021mlp,touvron2021resmlp,chen2021cyclemlp,yu2021s,guo2021hire,hou2021vision,yu2021rethinking} proposed various forms of spatial information (token) mixing/shuffling mechanisms to achieve competitive results without using explicit MSA operation.

\section{Super Token Transformer (STT)}
\label{method}
This section, first, describes the overall architecture of the proposed transformer. Then, we introduce the Super tokens and their use for modelling the local-to-global level associations. Finally, we describe the structure of our basic data encoder blocks and class learning blocks. 
\subsection{Overall Architecture}
The overall architecture of the proposed Super token transformer is illustrated in Figure~\ref{block-diagram}(a). First, an input image tokenization is performed using a four-layer convolutional stem with a patch size $\mathbf{P}$. We use convolution kernels of sizes (7,3,3,8) with strides (2,1,1.8) respectively. The resulting data tokens (DataTks) are partitioned into $M \times M$ sized windows. Next, a separate Super token (SupTk) is linked to each window resulting in $(M \times M)+1$ tokens per window. A learnable absolute positional embedding is then added to each token to learn the positional information. The prepared tokens are fed to the data encoder, formed by $L$ layers of the proposed data encoder blocks. Finally, the Suptk tokens and a CLS token are used to learn image representation using a two layered class embedding encoder.

\subsection{Super Tokens (SupTk)}
In a typical Transformer block, multi-head self-attention (MSA) is the most computationally expensive operation due to the pair-wise attention calculation among all input patches. Hence, local window based multi-head self-attention (WMSA) has been largely adopted in the recent works. In WMSA, self-attention is computed only across the tokens within each window. However, the best way of communicating information across windows has been an open question. 

The existing WMSA based solutions opted for a pyramid architecture, similar to CNN, with some mechanism of information seepage across neighbouring windows.  We now introduce our proposal of using window Super tokens. As mentioned earlier, we add one Super token to each local window. This learnable token is trained along with the other data tokens in a window through WMSA and feed-forward network. The function of the Super token is to learn a rich local embedding for a given window and then communicate this information across windows for global interaction modelling.

\subsection{Global Interaction with Super Token Mixer}
We model cross-window interactions on the global scale with the locally learnt Super tokens. A straight forward solution is to employ a transformer block with standard MSA. However, we found that the performance in terms of throughput and accuracy (c.f ablation study~\ref{ablation}) for this solution was poor. With the computational complexity and global scale in mind, we propose Super Token Mixer (STM) which uses separable convolution to interact across all windows. The global module first applies two depth-wise convolutions on Super tokens ($\mathbf{W}_{DW1},\:\mathbf{W}_{DW2}$), interchanging information across all Super tokens by operating on each channel separately. Next,  two point-wise convolutions ($\mathbf{W}_{PW1}, \mathbf{W}_{PW2}$) are applied to interact across all feature channels per spatial location. We use two residual connections across each convolutional block. Formally, the STM block can be written as:
\begin{center}
$y_{in} = [{SupTk_1}\; ||\; {SupTk_2}\; || \dots \; {SupTk_{N_s}} ]$
$y'\: =\: y_{in} + \: \mathbf{W}_{DW2} \: (GELU  (\mathbf{W}_{DW1}(\textbf{LN}(y_{in}))))$
$y_{out}\: =\: y' + \: \mathbf{W}_{PW2} \: (GELU  (\mathbf{W}_{PW1}(\textbf{LN}(y'))))$
\end{center}

The proposed global interaction module takes inspiration from the recent MLP-Mixer design~\cite{tolstikhin2021mlp}. All modern neural network architectures including CNNs, transformers and MLP based mixers have two main components to process information; channel wise information processing and spatial information processing. In typical CNN, this operation is defined by kernel size in the spatial domain, while taking into account all the channels. In separable CNNs, a 1x1 point-wise convolution is applied across the channels, while depth-wise convolution is applied across a group of channels in a local spatial neighbourhood. Vision transformers are essentially CNN with a slightly different way of processing spatial information. Tokenisation of the input image into data tokens is done by a convolutional layer (as the parameter are shared across all patches) but referred to as linear projection layer. Similarly layers to obtain the $Q, K$ and $V$ projections, multi-head information mixing layer in the MSA block, and the layers in FFN are all point-wise convolutions (as they share parameters across all the data tokens (features maps)), again renamed as so called linear layers. However, the spatial information mixing in a transformer is done by scaled dot-product attention between $Q$ and $K$ followed by softmax which serves as the coefficients/weights for $V$. This kind of spatial information mixing allows to capture long range relations. However, this leads to computational complexity and data/distillation requirements to train them (as there is no exploitation of local stationarity of the input signal in the design (locality bias), hence transformers require more data to learn natural locality biases present in images). 


\subsection{STT-Data Encoder Block}
Figure~\ref{block-diagram}(b) shows the two successive blocks of the proposed Super token transformer. First block consists of local windowed attention (WMSA) module and other standard modules (FFN, LayerNorm (\textbf{LN})). The subsequent block employs the proposed STM for mixing global information. Hence, forming an alternative local-global design. However, we demonstrate in the ablation study, that only the first global interaction is needed after learning the initial few layers locally. We adopted the per-channel LayerScale ($\lambda_{i}$) from \cite{touvron2021going} to dynamically scale each information channel. The mathematical formulation of the STT-data encoder at the layer $l$ is as follows:

\begin{center}
${x_l} = [{SupTk_l}\; ||\; {DataTk_l}]$

${x'} =  {x_l} + diag(\lambda_{l,1}, \dots, \lambda_{l,d} ) \times \textbf{WMSA}(\textbf{LN}(x_l))$

${x'} =  {x'} + diag(\lambda{'}_{l,1}, \dots, \lambda{'}_{l,d} ) \times \textbf{FFN}(\textbf{LN}(x'))$

${SupTks'} =\: \textbf{STM}\;({SupTks'})$

$x_{l+1} = [{SupTk'}\; ||\;  {DataTk'}]$
\end{center}

\subsection{STT-Class Embedding Block}
We employ a classification CLS token for the class embedding learning. We append the CLS token with the SupTk tokens in the last two layers of the network. We discard data patches entirely in the class embedding block and learn image class representation from SupTk tokens. We use standard MSA and FFN for modelling the final interaction among tokens. The CLS token is fed to the classifier head for class prediction. \begin{center}
${z_l} = [ CLS\; ||\; {SupTks} ]$

${z'} =  {z_l} + diag(\lambda_{l,1}, \dots, \lambda_{l,d} ) \times \textbf{MSA}(\textbf{LN}(z_l))$

${z_{l+1}} = {z'} + diag(\lambda{'}_{l,1}, \dots, \lambda{'}_{l,d}) \times \textbf{FFN}(\textbf{LN}(z'))$
\end{center}

\subsection{Complexity Analysis} 
The proposed Super tokens significantly reduce the memory consumption of the vision transformer. As mentioned earlier, the MSA is the most expensive operation with the $\mathcal{O}(N^2)$ complexity, where $N$ is the number of data tokens. With the WMSA and $M^2$ as the window size, the memory complexity reduces to $\mathcal{O}(NM^2)$. The complexity of the STM module is linear with $\mathcal{O}(N_s)$ where $N_s$ is the number of Super tokens. Finally, the class embedding layers employ MSA on Super Tokens and CLS token with the $\mathcal{O}(({N_s}+1)^2)$ complexity. Hence, with ${N_s}=16 << N=784$ in class layers, the overall complexity of the Super token transformer is $\sim \mathcal{O}(NM^2 + N_s)$ with $M=7$.

\bigskip
\noindent \textbf{Difference from Existing Approaches.}
The primary difference of the proposed STT architecture from the state-of-the-art methods is maintaining pure isotropic transformer design which is inclusive of WMSA, without requiring token merging. As mentioned in the introduction, interpretability of attention is compromised in pyramidal designs (Swin~\cite{liu2021swin}, Twins~\cite{chu2021twins}, PVT~\cite{wang2021pyramid}) due to strided convolution based token merging. In contrast, the proposed design provides a huge advantage in this regard, as the attention corresponding to the CLS token, which Super tokens are getting, can directly be passed down to local image tokens, giving more precise attentions (c.f.~\ref{visualize}). The MSG-Transformer~\cite{fang2021msg} has window messenger tokens that exchange information on regional level (group of nearby windows) by channel-groups shuffling. Thus, the purpose of a messenger token is only to exchange some feature channels with nearby windows. In contrast, the STT model exploits information on a global scale, and all feature channels of a Super token are involved in the flexible alternative local-global learning. Moreover, each Super token also maintains its identity and connection to the specific image window. RegionViT~\cite{chen2021regionvit} also exploits regional and local level interaction as the MSG-Transformer~\cite{fang2021msg}. However, different from the proposed STT, the regional tokens are coming from a parallel stream of transformer, having a bigger patch size than the local stream and initialised from data tokens.  

\section{Experimental Details}
\label{experiments}
\subsection{Architectural Variants}
The variants of the proposed transformer observe the following naming convention:  STT-S25, where the digits represent the number of layers $L$ in the STT-data encoder and the preceding letters represent the feature embedding dimension $C$. The hyper-parameter details of each variant are presented in the Table~\ref{variants}.

\begin{table}[]
\centering
\footnotesize{\caption{STT architecture variants. CLS token is learnt with extra two-layers on top of the layers mentioned below.}
\label{variants}}
\begin{tabular}{c|c|c|c}
\hline
STT-  & \# Layers & \# Heads & Dim \\ \hline
XXS25 & 25    & 4     & 192 \\
XXS37 & 37    & 4     & 192 \\
S25   & 25    & 8     & 384 \\
S37   & 37    & 8     & 384 \\
M25   & 25    & 16     & 768 \\
M37   & 37    & 16    & 768 \\ \hline
\end{tabular}
\end{table}

\subsection{Implementation Details}
We follow CaiT~\cite{touvron2021going} learning schedule to train our models. We use batch size 512 with base learning rate 0.0005 and five warm-up epochs. We adopt the AdamW~\cite{loshchilov2018fixing} optimizer with cosine learning rate scheduler. We apply Mixup~\cite{zhang2017mixup}, CutMix~\cite{yun2019cutmix}, RandomErasing~\cite{zhong2020random}, label smoothing~\cite{szegedy2016rethinking} and RandAugment~\cite{cubuk2020randaugment}. We show all results with image size $224 \times 224$, patch size 8 and $7 \times 7$ window size (data tokens per Super token). We use hard distillation loss~\cite{touvron2021training} and RegNetY-16GF~\cite{radosavovic2020designing} as teacher model for our ImageNet models. However, we perform fine-tuning on smaller datasets without distillation. We will submit our code in supplementary materials for complete details and release publicly upon acceptance. 

\noindent\textbf{Datasets.} We use ImageNet1K~\cite{deng2009imagenet} to train our main STT models. For fine-tuning, we use four downstream datasets including CIFAR-10~\cite{krizhevsky2009learning}, CIFAR-100~\cite{krizhevsky2009learning} and iNaturalist~\cite{van2018inaturalist} (iNat-2018, 2019).

\section{Results}
\label{results}
\subsection{Results on ImageNet-1K}
We compare the classification accuracy on ImageNet-1K benchmark in Table~\ref{imagenet-results}. Our small model STT-S25 with 384 embedding size and 49M parameters achieves 83.5\% classification accuracy. The results with bold font correspond to the designs comparable to the STT-S25 in terms of parameters. Under Isotrpic designs, DeiT-B~\cite{touvron2021training} suffers from high memory requirement although being only a 12-layer network. Our performance match with the CaiT-S24~\cite{touvron2021going} answers our research question by successfully encapsulating the information into the lower number of tokens (Super tokens) without degrading performance.

Further comparing to the window-based designs, the proposed model outperforms the corresponding Swin-S~\cite{liu2021swin}, MSGTr-S~\cite{fang2021msg} and RegionViT-M+~\cite{chen2021regionvit} in accuracy by using half the embedding size. In other words, we achieve on par accuracy performance as of the larger models (Swin-B, MSGTr-B, RegionViT-B etc) with our smaller model. We hope to get competitive results with our larger models as well which are still training.

\begin{table}[]
\centering
\footnotesize{\caption{Comparison between the SOTA methods on ImageNet-1K test set at $224 \times 224$ . All of the STT models are using distillation, patch size 8 and window size $7 \times 7$ for learning. \dagger represents distillation.The results for EfficientNet-B3~\cite{tan2019efficientnet} and ViT~\cite{dosovitskiy2020image} are reported on image size $300 \times 300$ and $384 \times 384$ respectively.}
\label{imagenet-results}}
\begin{threeparttable}
\resizebox{\linewidth}{!}{
\begin{tabular}{r|ccc|c}
\hline
\multirow{2}{*}{Network} & \#Params &  Throughput & Embedding  & ImgNet-1K \\  & (M)  & (Imgs/s) & Dimension   &  Top-1 (\%)  \\ \hline
RegNetY-16GF~\cite{radosavovic2020designing} & 84 & 334.7 & -   & 82.9 \\
EfficientNet-B3~\cite{tan2019efficientnet}  & 12 & 732.1 & -            & 81.6  \\ 
\specialrule{.1em}{.05em}{.05em}
 \multicolumn{5}{c}{\textbf{Isotropic Transformers}} \\ \specialrule{.1em}{.05em}{.05em}
ViT-B/16~\cite{dosovitskiy2020image}  & 86     & 85.9 & 768   & 77.9  \\
ViT-L/16~\cite{dosovitskiy2020image}    & 307    & 27.3 & 1024    & 76.5       \\ \hline
DeiT-Ti$^\dagger$~\cite{touvron2021training} & 6 &  2529.5 & 192   & 74.5 \\
DeiT-S$^\dagger$~\cite{touvron2021training} & 22 & 936.2 & 384 &  81.2\\
DeiT-B$^\dagger$~\cite{touvron2021training} & 87 & 290.9 & 768 &   83.4\\ \hline
CaiT-S24$^\dagger$~\cite{touvron2021going} & 47 & 573 & 384   & \textbf{83.5}\\
CaiT-S36~\cite{touvron2021going}  & 68 & 386 & 384  & 83.3\\
CaiT-S36$^\dagger$~\cite{touvron2021going} & 68 & - & 384  & 84.0 \\
CaiT-M24~\cite{touvron2021going}  & 186 & 262 & 768  & 83.4\\
CaiT-M24$^\dagger$~\cite{touvron2021going} & 186 & - & 768  & 84.7 \\\specialrule{.1em}{.05em}{.05em}
 \multicolumn{5}{c}{\textbf{Pyramidal Transformers}} \\ \specialrule{.1em}{.05em}{.05em}
Swin-T~\cite{liu2021swin} & 29 & 755.2  &  768     & 81.3 \\
Swin-S~\cite{liu2021swin} & 50 & 436.9 & 768   & \textbf{83.0}   \\
Swin-B~\cite{liu2021swin} & 88 & 278.1 & 1024  & 83.3 \\ \hline
MSGTr-T~\cite{fang2021msg} & 28  & 696.7 &  768  & 80.9 \\ 
MSGTr-S~\cite{fang2021msg} & 50  & 401.0 &  768  & \textbf{83.0} \\
MSGTr-B~\cite{fang2021msg} & 88  & 262.6 &  1024  & 83.5 \\ \hline
Twins-PCPVT-B~\cite{chu2021twins} & 44 & 525 &  512   & \textbf{82.7} \\
Twins-SVT-B~\cite{chu2021twins} & 56 &  469 &  768 & \textbf{83.2}\\ Twins-PCPVT-L~\cite{chu2021twins} & 60 & 367 &  512  & 83.1 \\ 
Twins-SVT-L~\cite{chu2021twins} & 99 &  288 &  1024   & 83.7\\\hline
NesT-T~\cite{zhang2021aggregating} & 17 & 633.9 & 192 & 81.5 \\
NesT-S~\cite{zhang2021aggregating} & 38 &  374.5 & 384  & \textbf{83.3}\\
NesT-B~\cite{zhang2021aggregating} & 68 &  235.8 & 768  & 83.8 \\ \hline
RegionViT-Ti+\cite{chen2021regionvit} &14 & - & 512  & 81.5\\ 
RegionViT-S+\cite{chen2021regionvit} &31 & - & 768  & 83.2\\
RegionViT-M+\cite{chen2021regionvit} & 42 & - & 768  & \textbf{83.4}\\
RegionViT-B\cite{chen2021regionvit} & 73 & - & 1024  & 83.3\\
RegionViT-B+\cite{chen2021regionvit} &74 & - & 1024  & 83.8 \\
\specialrule{.1em}{.05em}{.05em}
 \multicolumn{5}{c}{\textbf{Proposed Architecture}} \\ \specialrule{.1em}{.05em}{.05em}
STT-S25  & 49 & 735  & 384   & \textbf{83.5}  \\ 
STT-XXS25  & 12 & 602  & 192   & \textbf{80.7} \\ 
STT-XXS37  & 18 & -   & 192   & 79.00 (Epoch 283)\tnote{*}  \\ 
STT-S37  & 70 &  - & 384   & 82.02 (Epoch 288)\tnote{*}  \\ 
STT-M25  & 194 & - & 768  & 82.60 (Epoch 269)\tnote{*}  \\
STT-M37  & 279 & - & 768  & 76.06 (Epoch 63)\tnote{*}  \\
\hline
\end{tabular}}
\flushleft  \footnotesize{* Training process is still in progress. Updated results will be provided in the supplementary materials.}
\end{threeparttable}
\end{table}

\subsection{Results with Fine-tuning on Small Datasets}
In Table ~\ref{finetuning}, We present results for fine-tuning the STT-S25 on smaller sized datasets. Different works have used various evaluation protocols for the downstream evaluation (e.g. 384 or 32 as image size, number of epochs for fine-tuning, crop-size etc),  making it hard to compare fairly. In the table, we are comparing with methods providing results at $224 \times 224$. Interestingly, The STT-S25 model outperforms the DeiT-B~\cite{touvron2021training} on high-resolution image sets (iNat-2018 and 2019) but accuracy drops on the low-resolution CIFAR data. We leave its further investigation to future work when we will get complete results for our other models as well.      

\begin{table}[]
\centering
\footnotesize{\caption{The results of fine-tuning the pretrained STT-S25 on downstream datasets at resolution $224 \times 224$.}
\label{finetuning}}
\begin{threeparttable}
\resizebox{\linewidth}{!}{
\begin{tabular}{r|c|c|c|c}
\hline
Model & CIFAR10   & CIFAR100 & iNat2018  & iNat2019 \\ \hline
EfficientNet-B7~\cite{tan2019efficientnet} & 98.9 & 91.7 & - & -\\
ViT-B/16~\cite{dosovitskiy2020image} & 98.1 & 87.1 & -& -\\
ViT-L/16~\cite{dosovitskiy2020image} & 97.9 & 86.4 &- &- \\
Deit-B~\cite{touvron2021training} & 99.1 & 90.8 & 73.2  & 77.7\\
STT-S25 & 97.77\tnote{*} & 83.00\tnote{*} & 74.59 & 77.89 \\ \hline
\end{tabular}}
\flushleft  \footnotesize{* Training in progress. Updated results will be provided in the supplementary materials.}
\end{threeparttable}
\end{table}

\section{Ablation Study}
\label{ablation}
In this section, we present the ablation analysis for the proposed method. Due to limited computational capacity, we perform ablation study on randomly selected 50 classes from ImageNet-1K, termed as ImageNet-50. We used STT-S25 model trained for 300 epochs for the ablation study and kept all hyper-parameters identical to those used in the main experiment unless specified.

\begin{table}[]
\centering
\footnotesize{\caption{Ablation study for global interaction designs on ImageNet-50}
\label{global-design}}
\begin{tabular}{c|c|c|c}
\hline
ImageNet-50 & \#Params (M)   & Throughput (Imgs/s)  & Top-1 Acc (\%) \\ \hline
MSA         & 46.97   & 416 & 87.84    \\ 
Conv (2x2)  & 46.96  & 1250 &  84.80   \\
Conv (3x3) & 54.33  & 2500 & 85.44 \\
STM  & 47.04  & 833 & 89.20 \\ \hline
\end{tabular}
\label{cross-layers}
\end{table}

\subsection{Design Choices for Global Interaction}
We investigated various schemes for global interaction in Table~\ref{global-design}. As mentioned earlier, we tested with standard transformer block for global information consolidation and got 1.36\% lower accuracy than the proposed STM module. We observed lower throughput as well.We suspect that transformer block does have weak locality bias in the sense that the correlation between transformed views (Query and Key) corresponding to the same token and tokens around this token will have stronger correlations. Hence, the nearby tokens will have higher weights for V in dot product attention. This is evident from the stronger block diagonal lines in attention visualisation of any transformers, particularly in initial blocks. This weak locality bias leads to favouring more contribution from the token itself, which may not be the most suitable mechanism to consolidate global information in our case. We also investigated 2$D$ convolution kernels, however, the performance dropped even below the MSA. We believe that it created the effect of regional interaction instead of global interaction and overlooked critical information.

\subsection{Granularity of Super Tokens}
We performed an experiment to study the effect of the number of windows/Super tokens. The results are presented in Table~\ref{super-tokens-granularity}. It is critical to understand this behaviour for optimal global representation learning. Ignoring the computational complexity, it is obvious that using more Super tokens would help in modelling global interaction. However, it is interesting to find that the performance decreased by 0.76\% when we increased the number of windows to 49 from 16. One reason could be that by increasing windows, more redundant information is seeping to the global layers, affecting the discriminatory power of the transformer. On the other hand, we lose a lot of local details, which are critical to the accuracy of prediction, by decreasing the number of Super tokens (1.48\% accuracy drop). Although, now the Super token has a larger receptive field,  one token is not enough for summarising important local details. It would be interesting to investigate the effect of increasing  Super token depth versus number of windows.  

\begin{table}[]
\centering
\footnotesize {\caption{Effect of number of Super tokens and window size on ImageNet-50 with the STT-25 model}
\label{super-tokens-granularity}}
\begin{tabular}{c|c|c}
\hline
\#SupTks ($N_s$)  & Window Size ($M \times M$) & Top-1 Acc(\%) \\ \hline
4 & 14$\times$14    & 87.72     \\
16 & 7$\times$7    & 89.20     \\
49  & 4$\times$4  & 88.44     \\ \hline
\end{tabular}
\end{table}

\subsection{Position of the STM Module}
We studied the impact of global interaction starting at different layers. The corresponding results are provided in Table~\ref{STM-ablation}. Since Super tokens are randomly initialised, the earlier interaction (layer 2,4) does not perform well indicating the need for learning richer local representation. Interacting from higher layers (layer 10,12,18) suggests losing details from the lower layers that are critical for the cross-window communication. In our experiments, we adopted layer 6 as the starting position for global interaction as it provided the optimal point for maintaining cross-windows links and obtaining rich enough local window representations. Hence, first five layers of the STT-data encoder perform pure local level learning and the following layers follow alternate local-global learning pattern.   

\begin{table}[]
\centering
\footnotesize{\caption{Ablation study for the position of the global STM module}
\label{STM-ablation}}
\begin{tabular}{c|c|c}
\hline
Layer & \#Params (M)    &  Top-1 Acc (\%) \\ \hline
2 & 47.06  & 88.56 \\
4 & 47.05    & 88.56  \\
6 &  47.04    & 89.20 \\
8 & 47.03  & 89.16 \\
10 & 47.03 & 88.28  \\
12 & 47.02   & 88.92 \\
18 & 47.00  & 88.76 \\ \hline
\end{tabular}
\label{cross-layers}
\end{table}

\begin{figure*}[]
    \centering
    \begin{subfigure}[t]{0.4\textwidth}
        \centering
        \caption*{conf: 0.83 - pred: hen - GT: cock}
        \includegraphics[width=0.32\textwidth]{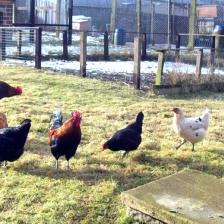}
        \includegraphics[width=0.32\textwidth]{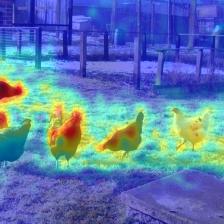}
        \includegraphics[width=0.32\textwidth]{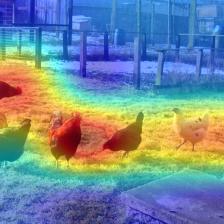}
    \end{subfigure}
    \hspace{0.5cm}
    \begin{subfigure}[t]{0.4\textwidth}
        \centering
        \caption*{conf: 0.95 - pred: mousetrap - GT: mousetrap}
        \includegraphics[width=0.32\textwidth]{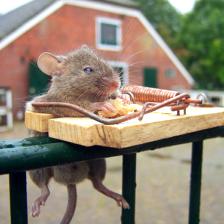}
        \includegraphics[width=0.32\textwidth]{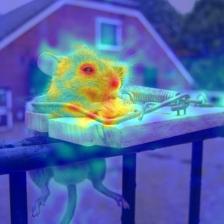}
        \includegraphics[width=0.32\textwidth]{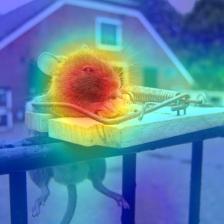}
    \end{subfigure}
    
    \vspace{0.3cm}
    \begin{subfigure}[t]{0.4\textwidth}
        \centering
        \caption*{conf: 0.91 - pred: hedgehog - GT: hedgehog}
        \includegraphics[width=0.32\textwidth]{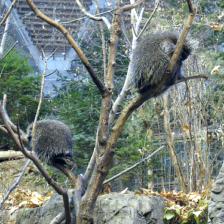}
        \includegraphics[width=0.32\textwidth]{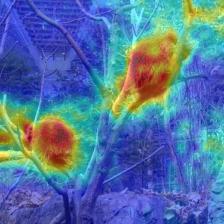}
        \includegraphics[width=0.32\textwidth]{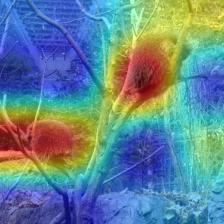}
    \end{subfigure}%
    \hspace{0.5cm}
    \begin{subfigure}[t]{0.4\textwidth}
        \centering
        \caption*{conf: 0.93 - pred: c. butterfly - GT: c. butterfly}
        \includegraphics[width=0.32\textwidth]{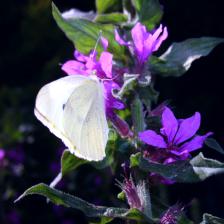}
        \includegraphics[width=0.32\textwidth]{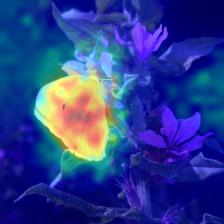}
        \includegraphics[width=0.32\textwidth]{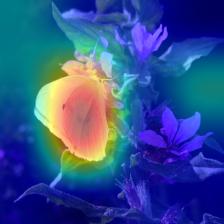}
    \end{subfigure}
        
    \vspace{0.3cm}
    \begin{subfigure}[t]{0.4\textwidth}
        \centering
        \caption*{conf: 0.86 - pred: sea lion - GT: sea lion}
        \includegraphics[width=0.32\textwidth]{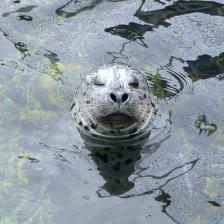}
        \includegraphics[width=0.32\textwidth]{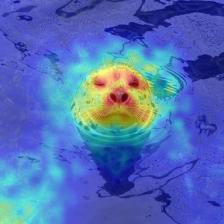}
        \includegraphics[width=0.32\textwidth]{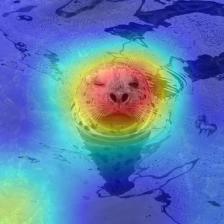}
    \end{subfigure}%
    \hspace{0.5cm}
    \begin{subfigure}[t]{0.4\textwidth}
        \centering
        \caption*{conf: 0.53 - pred: tripod - GT: tripod}
        \includegraphics[width=0.32\textwidth]{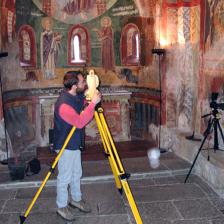}
        \includegraphics[width=0.32\textwidth]{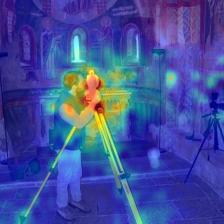}
        \includegraphics[width=0.32\textwidth]{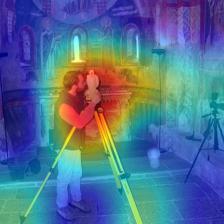}
    \end{subfigure}
    
    \vspace{0.3cm}
    \begin{subfigure}[t]{0.4\textwidth}
        \centering
        \caption*{conf: 0.91 - pred: tricycle - GT: tricycle}
        \includegraphics[width=0.32\textwidth]{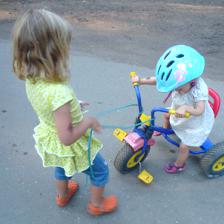}
        \includegraphics[width=0.32\textwidth]{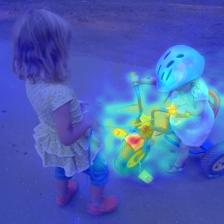}
        \includegraphics[width=0.32\textwidth]{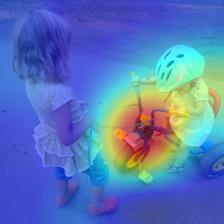}
    \end{subfigure}%
    \hspace{0.5cm}
    \begin{subfigure}[t]{0.4\textwidth}
        \centering
        \caption*{conf: 0.95 - pred: kite - GT: kite}
        \includegraphics[width=0.32\textwidth]{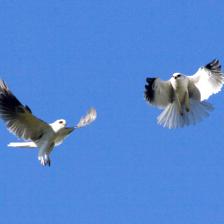}
        \includegraphics[width=0.32\textwidth]{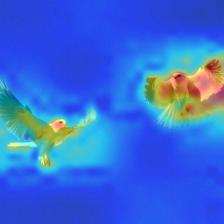}
        \includegraphics[width=0.32\textwidth]{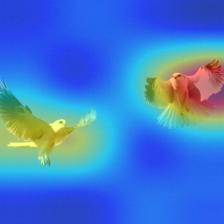}
    \end{subfigure}
        
    \vspace{0.3cm}
    \begin{subfigure}[t]{0.4\textwidth}
        \centering
        \caption*{conf: 0.91 - pred: A. lobster - GT: A. lobster }
        \includegraphics[width=0.32\textwidth]{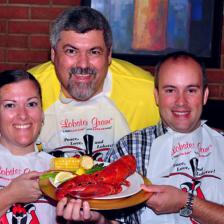}
        \includegraphics[width=0.32\textwidth]{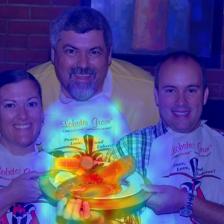}
        \includegraphics[width=0.32\textwidth]{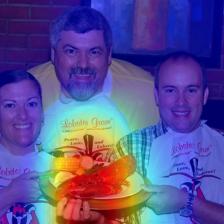}
    \end{subfigure}%
    \hspace{0.5cm}
    \begin{subfigure}[t]{0.4\textwidth}
        \centering
        \caption*{conf: 0.91 - pred: impala - GT: impala}
        \includegraphics[width=0.32\textwidth]{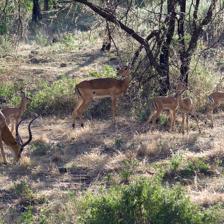}
        \includegraphics[width=0.32\textwidth]{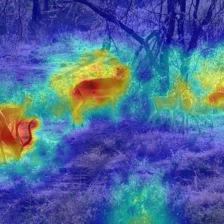}
        \includegraphics[width=0.32\textwidth]{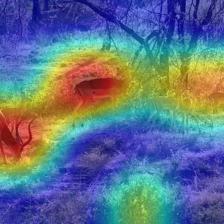}
    \end{subfigure}
        
    \vspace{0.3cm}
    \begin{subfigure}[t]{0.4\textwidth}
        \centering
        \caption*{conf: 0.92 - pred: snowmobile - GT: snowmobile}
        \includegraphics[width=0.32\textwidth]{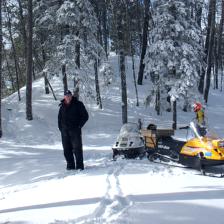}
        \includegraphics[width=0.32\textwidth]{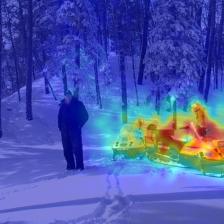}
        \includegraphics[width=0.32\textwidth]{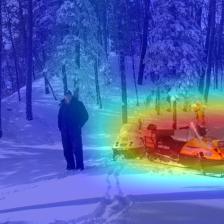}
    \end{subfigure}%
    \hspace{0.5cm}
    \begin{subfigure}[t]{0.4\textwidth}
        \centering
        \caption*{conf: 0.85 - pred: carriage dog - GT: flute}
        \includegraphics[width=0.32\textwidth]{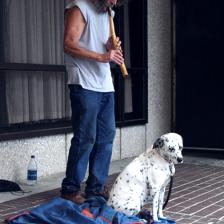}
        \includegraphics[width=0.32\textwidth]{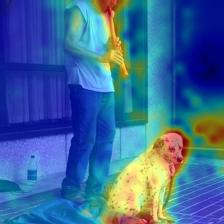}
        \includegraphics[width=0.32\textwidth]{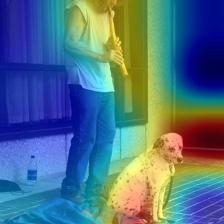}
    \end{subfigure}
    
    \vspace{0.3cm}
    \begin{subfigure}[t]{0.4\textwidth}
        \centering
        \caption*{conf: 0.92 - pred: skunk - GT: skunk}
        \includegraphics[width=0.32\textwidth]{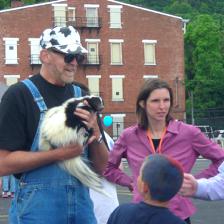}
        \includegraphics[width=0.32\textwidth]{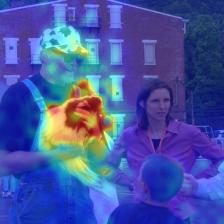}
        \includegraphics[width=0.32\textwidth]{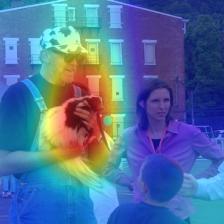}
    \end{subfigure}%
    \hspace{0.5cm} 
    \begin{subfigure}[t]{0.4\textwidth}
        \centering
        \caption*{conf: 0.93 - pred: bulbul - GT: bulbul}
        \includegraphics[width=0.32\textwidth]{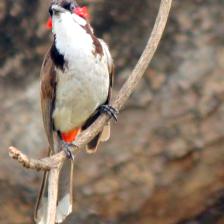}
        \includegraphics[width=0.32\textwidth]{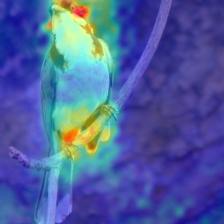}
        \includegraphics[width=0.32\textwidth]{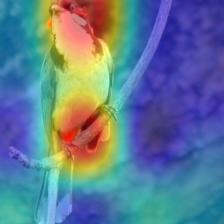}
    \end{subfigure}
    \caption{Attention visualisation with STT-S25 model. Images are arranged in three columns. \textit{left:} input image, \textit{middle:} local window attentions from the STT-data encoder (\textbf{Layer 25}), \textit{right:} global Super tokens attentions from the STT-class embedding block (\textbf{Layer 26}).
    \label{visualize}}
\end{figure*}

\subsection{Attention Visualisations} 
In Figure~\ref{visualize}, we provide the average attention maps across the 8 heads of the STT-S25 model trained on ImageNet-1K. Images are arranged in three columns as original, window attentions from Layer 25 and Super tokens attentions from Layer 26. Taking the mouse trap class from the top row as an example, we can note that Super tokens provide a blobby attention to most important image windows for the object. However, the window attention pays more focus on discriminative features of the class (eyes, nose, tail of the mouse). It is encouraging to witness that the proposed Super tokens successfully encapsulate the image content into a lower number of tokens.

\section{Discussion} 
Regarding our research aims for this work, we found that our hypothesis to encapsulate information into a lower number of tokens and then use these tokens for decision making is successfully validated by the experimental results. Moreover, with the ablation study, we reckon that the proposed design is flexible and adaptable to different architectural configurations. This property, together with the ability of supporting fine-grained attention on the local tokens, makes the STT architecture a strong candidate for dense prediction tasks like localisation and segmentation, while being efficient. The power and compute considerations are becoming critical as the neural network models are getting bigger. The proposed efficient solution for local and global information handling enables our design to be significantly faster than vanilla transformers, reducing carbon footprint and leading to the responsible AI.

\subsection{Limitations}
As noted in  Section~\ref{results}, we plan to investigate the effect of fine-tuning the Super tokens on various image resolutions. Moreover, further studies are required for the efficient Super token exploitation in the case of images containing confusing/multiple classes. 

\bigskip
\noindent\textbf{Resource bias, diversity and inclusivity statement.} We demonstrate most of the results on a randomly selected 50 class subset of the ImageNet-1K dataset, due to the fact that the research is coming from a small group with only 1.6 GPU per researcher, rather than having access to thousands of high capacity GPUs and a multi-million dollar budget for one training. We hope the proposed design will not be judged through typical status-quo glasses. Our hope is that the inability to perform large scale experiments will not be used to penalise less resourced groups, but instead, the focus will be on the scientific contribution, and its demonstrated merits, as well as its potential impact on large scale deep learning problems.

\section{Conclusion}
In this paper, we proposed a new design for vision transformer based on local to global interaction modelling through special tokens, referred as Super tokens. The design exploits a local windowed MSA, without an information degrading token merging operation, and maintains a typical isotropic transformer structure. The proposed Super tokens simultaneously preserve the detailed local information and communicate it globally to model cross-window dependencies. Moreover, the design offers computational benefits with competitive experimental performance.

{\small
\bibliographystyle{ieee_fullname}
\bibliography{egbib}

\begin{thebibliography}{10}\itemsep=-1pt

\bibitem{atito2021sit}
Sara Atito, Muhammad Awais, and Josef Kittler.
\newblock Sit: Self-supervised vision transformer.
\newblock {\em arXiv preprint arXiv:2104.03602}, 2021.

\bibitem{caron2021emerging}
Mathilde Caron, Hugo Touvron, Ishan Misra, Herv{\'e} J{\'e}gou, Julien Mairal,
  Piotr Bojanowski, and Armand Joulin.
\newblock Emerging properties in self-supervised vision transformers.
\newblock {\em arXiv preprint arXiv:2104.14294}, 2021.

\bibitem{chen2021regionvit}
Chun-Fu Chen, Rameswar Panda, and Quanfu Fan.
\newblock Regionvit: Regional-to-local attention for vision transformers.
\newblock {\em arXiv preprint arXiv:2106.02689}, 2021.

\bibitem{chen2021cyclemlp}
Shoufa Chen, Enze Xie, Chongjian Ge, Ding Liang, and Ping Luo.
\newblock Cyclemlp: A mlp-like architecture for dense prediction.
\newblock {\em arXiv preprint arXiv:2107.10224}, 2021.

\bibitem{chu2021twins}
Xiangxiang Chu, Zhi Tian, Yuqing Wang, Bo Zhang, Haibing Ren, Xiaolin Wei,
  Huaxia Xia, and Chunhua Shen.
\newblock Twins: Revisiting the design of spatial attention in vision
  transformers.
\newblock {\em arXiv preprint arXiv:2104.13840}, 1(2):3, 2021.

\bibitem{cubuk2020randaugment}
Ekin~D Cubuk, Barret Zoph, Jonathon Shlens, and Quoc~V Le.
\newblock Randaugment: Practical automated data augmentation with a reduced
  search space.
\newblock In {\em Proceedings of the IEEE/CVF Conference on Computer Vision and
  Pattern Recognition Workshops}, pages 702--703, 2020.

\bibitem{deng2009imagenet}
Jia Deng, Wei Dong, Richard Socher, Li-Jia Li, Kai Li, and Li Fei-Fei.
\newblock Imagenet: A large-scale hierarchical image database.
\newblock In {\em 2009 IEEE conference on computer vision and pattern
  recognition}, pages 248--255. Ieee, 2009.

\bibitem{devlin2018bert}
Jacob Devlin, Ming-Wei Chang, Kenton Lee, and Kristina Toutanova.
\newblock Bert: Pre-training of deep bidirectional transformers for language
  understanding.
\newblock {\em arXiv preprint arXiv:1810.04805}, 2018.

\bibitem{dong2018speech}
Linhao Dong, Shuang Xu, and Bo Xu.
\newblock Speech-transformer: a no-recurrence sequence-to-sequence model for
  speech recognition.
\newblock In {\em 2018 IEEE International Conference on Acoustics, Speech and
  Signal Processing (ICASSP)}, pages 5884--5888. IEEE, 2018.

\bibitem{dong2021cswin}
Xiaoyi Dong, Jianmin Bao, Dongdong Chen, Weiming Zhang, Nenghai Yu, Lu Yuan,
  Dong Chen, and Baining Guo.
\newblock Cswin transformer: A general vision transformer backbone with
  cross-shaped windows.
\newblock {\em arXiv preprint arXiv:2107.00652}, 2021.

\bibitem{dosovitskiy2020image}
Alexey Dosovitskiy, Lucas Beyer, Alexander Kolesnikov, Dirk Weissenborn,
  Xiaohua Zhai, Thomas Unterthiner, Mostafa Dehghani, Matthias Minderer, Georg
  Heigold, Sylvain Gelly, et~al.
\newblock An image is worth 16x16 words: Transformers for image recognition at
  scale.
\newblock {\em arXiv preprint arXiv:2010.11929}, 2020.

\bibitem{el2021xcit}
Alaaeldin El-Nouby, Hugo Touvron, Mathilde Caron, Piotr Bojanowski, Matthijs
  Douze, Armand Joulin, Ivan Laptev, Natalia Neverova, Gabriel Synnaeve, Jakob
  Verbeek, et~al.
\newblock Xcit: Cross-covariance image transformers.
\newblock {\em arXiv preprint arXiv:2106.09681}, 2021.

\bibitem{fang2021msg}
Jiemin Fang, Lingxi Xie, Xinggang Wang, Xiaopeng Zhang, Wenyu Liu, and Qi Tian.
\newblock Msg-transformer: Exchanging local spatial information by manipulating
  messenger tokens.
\newblock {\em arXiv preprint arXiv:2105.15168}, 2021.

\bibitem{gulati2020conformer}
Anmol Gulati, James Qin, Chung-Cheng Chiu, Niki Parmar, Yu Zhang, Jiahui Yu,
  Wei Han, Shibo Wang, Zhengdong Zhang, Yonghui Wu, et~al.
\newblock Conformer: Convolution-augmented transformer for speech recognition.
\newblock {\em arXiv preprint arXiv:2005.08100}, 2020.

\bibitem{guo2021hire}
Jianyuan Guo, Yehui Tang, Kai Han, Xinghao Chen, Han Wu, Chao Xu, Chang Xu, and
  Yunhe Wang.
\newblock Hire-mlp: Vision mlp via hierarchical rearrangement.
\newblock {\em arXiv preprint arXiv:2108.13341}, 2021.

\bibitem{han2021transformer}
Kai Han, An Xiao, Enhua Wu, Jianyuan Guo, Chunjing Xu, and Yunhe Wang.
\newblock Transformer in transformer.
\newblock {\em arXiv preprint arXiv:2103.00112}, 2021.

\bibitem{hou2021vision}
Qibin Hou, Zihang Jiang, Li Yuan, Ming-Ming Cheng, Shuicheng Yan, and Jiashi
  Feng.
\newblock Vision permutator: A permutable mlp-like architecture for visual
  recognition.
\newblock {\em arXiv preprint arXiv:2106.12368}, 2021.

\bibitem{kitaev2020reformer}
Nikita Kitaev, {\L}ukasz Kaiser, and Anselm Levskaya.
\newblock Reformer: The efficient transformer.
\newblock {\em arXiv preprint arXiv:2001.04451}, 2020.

\bibitem{krizhevsky2009learning}
Alex Krizhevsky, Geoffrey Hinton, et~al.
\newblock Learning multiple layers of features from tiny images.
\newblock 2009.

\bibitem{li2020oscar}
Xiujun Li, Xi Yin, Chunyuan Li, Pengchuan Zhang, Xiaowei Hu, Lei Zhang, Lijuan
  Wang, Houdong Hu, Li Dong, Furu Wei, et~al.
\newblock Oscar: Object-semantics aligned pre-training for vision-language
  tasks.
\newblock In {\em European Conference on Computer Vision}, pages 121--137.
  Springer, 2020.

\bibitem{liu2021swin}
Ze Liu, Yutong Lin, Yue Cao, Han Hu, Yixuan Wei, Zheng Zhang, Stephen Lin, and
  Baining Guo.
\newblock Swin transformer: Hierarchical vision transformer using shifted
  windows.
\newblock {\em arXiv preprint arXiv:2103.14030}, 2021.

\bibitem{loshchilov2018fixing}
Ilya Loshchilov and Frank Hutter.
\newblock Fixing weight decay regularization in adam.
\newblock 2018.

\bibitem{radosavovic2020designing}
Ilija Radosavovic, Raj~Prateek Kosaraju, Ross Girshick, Kaiming He, and Piotr
  Doll{\'a}r.
\newblock Designing network design spaces.
\newblock In {\em Proceedings of the IEEE/CVF Conference on Computer Vision and
  Pattern Recognition}, pages 10428--10436, 2020.

\bibitem{szegedy2016rethinking}
Christian Szegedy, Vincent Vanhoucke, Sergey Ioffe, Jon Shlens, and Zbigniew
  Wojna.
\newblock Rethinking the inception architecture for computer vision.
\newblock In {\em Proceedings of the IEEE conference on computer vision and
  pattern recognition}, pages 2818--2826, 2016.

\bibitem{tan2019efficientnet}
Mingxing Tan and Quoc Le.
\newblock Efficientnet: Rethinking model scaling for convolutional neural
  networks.
\newblock In {\em International Conference on Machine Learning}, pages
  6105--6114. PMLR, 2019.

\bibitem{tolstikhin2021mlp}
Ilya Tolstikhin, Neil Houlsby, Alexander Kolesnikov, Lucas Beyer, Xiaohua Zhai,
  Thomas Unterthiner, Jessica Yung, Andreas Steiner, Daniel Keysers, Jakob
  Uszkoreit, et~al.
\newblock Mlp-mixer: An all-mlp architecture for vision.
\newblock {\em arXiv preprint arXiv:2105.01601}, 2021.

\bibitem{touvron2021resmlp}
Hugo Touvron, Piotr Bojanowski, Mathilde Caron, Matthieu Cord, Alaaeldin
  El-Nouby, Edouard Grave, Gautier Izacard, Armand Joulin, Gabriel Synnaeve,
  Jakob Verbeek, et~al.
\newblock Resmlp: Feedforward networks for image classification with
  data-efficient training.
\newblock {\em arXiv preprint arXiv:2105.03404}, 2021.

\bibitem{touvron2021training}
Hugo Touvron, Matthieu Cord, Matthijs Douze, Francisco Massa, Alexandre
  Sablayrolles, and Herv{\'e} J{\'e}gou.
\newblock Training data-efficient image transformers \& distillation through
  attention.
\newblock In {\em International Conference on Machine Learning}, pages
  10347--10357. PMLR, 2021.

\bibitem{touvron2021going}
Hugo Touvron, Matthieu Cord, Alexandre Sablayrolles, Gabriel Synnaeve, and
  Herv{\'e} J{\'e}gou.
\newblock Going deeper with image transformers.
\newblock {\em arXiv preprint arXiv:2103.17239}, 2021.

\bibitem{van2018inaturalist}
Grant Van~Horn, Oisin Mac~Aodha, Yang Song, Yin Cui, Chen Sun, Alex Shepard,
  Hartwig Adam, Pietro Perona, and Serge Belongie.
\newblock The inaturalist species classification and detection dataset.
\newblock In {\em Proceedings of the IEEE conference on computer vision and
  pattern recognition}, pages 8769--8778, 2018.

\bibitem{vaswani2017attention}
Ashish Vaswani, Noam Shazeer, Niki Parmar, Jakob Uszkoreit, Llion Jones,
  Aidan~N Gomez, {\L}ukasz Kaiser, and Illia Polosukhin.
\newblock Attention is all you need.
\newblock In {\em Advances in neural information processing systems}, pages
  5998--6008, 2017.

\bibitem{wang2021pyramid}
Wenhai Wang, Enze Xie, Xiang Li, Deng-Ping Fan, Kaitao Song, Ding Liang, Tong
  Lu, Ping Luo, and Ling Shao.
\newblock Pyramid vision transformer: A versatile backbone for dense prediction
  without convolutions.
\newblock {\em arXiv preprint arXiv:2102.12122}, 2021.

\bibitem{yu2021rethinking}
Tan Yu, Xu Li, Yunfeng Cai, Mingming Sun, and Ping Li.
\newblock Rethinking token-mixing mlp for mlp-based vision backbone.
\newblock {\em arXiv preprint arXiv:2106.14882}, 2021.

\bibitem{yu2021s}
Tan Yu, Xu Li, Yunfeng Cai, Mingming Sun, and Ping Li.
\newblock $s^{2}$-mlp: Spatial-shift mlp architecture for vision.
\newblock {\em arXiv preprint arXiv:2106.07477}, 2021.

\bibitem{yuan2021tokens}
Li Yuan, Yunpeng Chen, Tao Wang, Weihao Yu, Yujun Shi, Zihang Jiang, Francis~EH
  Tay, Jiashi Feng, and Shuicheng Yan.
\newblock Tokens-to-token vit: Training vision transformers from scratch on
  imagenet.
\newblock {\em arXiv preprint arXiv:2101.11986}, 2021.

\bibitem{yun2019cutmix}
Sangdoo Yun, Dongyoon Han, Seong~Joon Oh, Sanghyuk Chun, Junsuk Choe, and
  Youngjoon Yoo.
\newblock Cutmix: Regularization strategy to train strong classifiers with
  localizable features.
\newblock In {\em Proceedings of the IEEE/CVF International Conference on
  Computer Vision}, pages 6023--6032, 2019.

\bibitem{zhang2017mixup}
Hongyi Zhang, Moustapha Cisse, Yann~N Dauphin, and David Lopez-Paz.
\newblock mixup: Beyond empirical risk minimization.
\newblock {\em arXiv preprint arXiv:1710.09412}, 2017.

\bibitem{zhang2021aggregating}
Zizhao Zhang, Han Zhang, Long Zhao, Ting Chen, and Tomas Pfister.
\newblock Aggregating nested transformers.
\newblock {\em arXiv preprint arXiv:2105.12723}, 2021.

\bibitem{zhong2020random}
Zhun Zhong, Liang Zheng, Guoliang Kang, Shaozi Li, and Yi Yang.
\newblock Random erasing data augmentation.
\newblock In {\em Proceedings of the AAAI Conference on Artificial
  Intelligence}, volume~34, pages 13001--13008, 2020.

\end{thebibliography}
}

\end{document}